\documentclass[sigconf]{acmart}

\AtBeginDocument{%
	\providecommand\BibTeX{{%
			\normalfont B\kern-0.5em{\scshape i\kern-0.25em b}\kern-0.8em\TeX}}}
		
\usepackage{amsmath,amsfonts}
\usepackage{graphicx}
\usepackage{textcomp}
\usepackage{xcolor}

\usepackage{subfigure}

\usepackage{diagbox}
\usepackage{threeparttable}
\usepackage{multirow}
\usepackage{color}

\usepackage{url}
\usepackage{bbold}

\usepackage{setspace}

\usepackage{comment}

\usepackage{balance}

\usepackage{algpseudocode}
\usepackage[ruled]{algorithm2e}

\usepackage{colortbl}
\usepackage{graphicx}
\usepackage{makecell}
\usepackage{wrapfig}
\usepackage{url}

\usepackage{bbold}
\usepackage{enumitem}

\copyrightyear{2024}
\acmYear{2024}
\setcopyright{acmlicensed}
\acmConference[CIKM '24] {Proceedings of the 33rd ACM International Conference on Information and Knowledge Management}{October 21--25, 2024}{Boise, ID, USA.}
\acmBooktitle{Proceedings of the 33rd ACM International Conference on Information and Knowledge Management (CIKM '24), October 21--25, 2024, Boise, ID, USA}
\acmISBN{979-8-4007-0436-9/24/10}
\acmDOI{10.1145/XXXXXX.XXXXXX}

\begin{document}


\title{Fairness in Large Language Models in Three Hours}



\author{Thang Viet Doan}
\orcid{0009-0009-3072-5532}
\affiliation{%
  \institution{Florida International University}
  \city{Miami}
  \state{FL}
  \country{US}}
\email{tdoan011@fiu.edu}

\author{Zichong Wang}
\orcid{0000-0001-6091-6609}
\affiliation{%
  \institution{Florida International University}
  \city{Miami}
  \state{FL}
  \country{US}}
\email{ziwang@fiu.edu}

\author{Minh Nhat Nguyen}
\orcid{0009-0005-7539-5259}
\affiliation{%
  \institution{Florida International University}
  \city{Miami}
  \state{FL}
  \country{US}}
\email{nhoan009@fiu.edu}

\author{Wenbin Zhang}
\orcid{0000-0003-3024-5415}
\affiliation{%
  \institution{Florida International University}
  \city{Miami}
  \state{FL}
  \country{US}}
\email{wenbin.zhang@fiu.edu}
\authornote{Corresponding author}






\renewcommand{\shortauthors}{Thang Doan Viet, Zichong Wang, Minh Nhat Nguyen, \& Wenbin Zhang}

\begin{abstract}

Large Language Models (LLMs) have demonstrated remarkable success across various domains but often lack fairness considerations, potentially leading to discriminatory outcomes against marginalized populations. Unlike fairness in traditional machine learning, fairness in LLMs involves unique backgrounds, taxonomies, and fulfillment techniques. This tutorial provides a systematic overview of recent advances in the literature concerning fair LLMs, beginning with real-world case studies to introduce LLMs, followed by an analysis of bias causes therein. The concept of fairness in LLMs is then explored, summarizing the strategies for evaluating bias and the algorithms designed to promote fairness. Additionally, resources for assessing bias in LLMs, including toolkits and datasets, are compiled, and current research challenges and open questions in the field are discussed.  The repository is available on this \href{https://github.com/LavinWong/Fairness-in-Large-Language-Models}{website}\footnote{\url{https://github.com/LavinWong/Fairness-in-Large-Language-Models}}.
\end{abstract}




\maketitle


\section{Introduction} 

Large Language Models (LLMs), such as BERT~\cite{devlin2018bert}, GPT-3~\cite{brown2020language}, and LLaMA~\cite{touvron2023llama}, have shown powerful performance and development prospects in various tasks of Natural Language Processing due to their robust text encoding and decoding capabilities and discovered emergent capabilities (\textit{e.g.,} reasoning)~\cite{chu2024history}. Despite their great performance, LLMs tend to inherit bias from multiple sources, including training data, encoding processes, and fine-tuning procedures, which may result in biased decisions against certain groups defined by the \textit{sensitive attribute} (\textit{e.g.,} age, gender, or race).
The biased prediction has raised significant ethical and societal concerns, severely limiting the adoption of LLMs in high-risk decision-making scenarios such as hiring,
loan approvals,
legal sentencing,
and medical diagnoses.

To this end, many efforts have been made to mitigate bias in LLMs~\cite{kotek2023gender, yogarajan2023tackling, zhang2023chatgpt}. For example, one line of work extends traditional fairness notions—individual fairness and group fairness—to these models~\cite{chhikara2024few}. Specifically, individual fairness seeks to ensure similar outcomes for similar individuals~\cite{dwork2012fairness,zhang2023individual}, while group fairness focuses on equalizing outcome statistics across subgroups defined by sensitive attributes~\cite{hardt2016equality,wang2023preventing,wang2023fairness,wang2023mitigating} (\textit{e.g.,} gender or race). While these classification-based fairness notions are adept at evaluating bias in LLM's classification results~\cite{chhikara2024few}, they fall short in addressing biases that arise during the LLM generation process
~\cite{hu2024unveiling}. In other words, LLMs demand a nuanced approach to measure and mitigate bias that emerges both in their outputs and during the generation process. This complexity motivates other lines of linguistic strategies that not only evaluate the accuracy of LLMs but also their propagation of harmful stereotypes or discriminatory language. For instance, a study  
examining the behavior of an LLM like ChatGPT revealed a concerning trend: it generated letters of recommendation that described a fictitious individual named Kelly (\textit{i.e.,} a commonly female-associated name) as ``warm and amiable'', while describing Joseph (\textit{i.e.,} a commonly male-associated name) as a ``natural leader and role model''. This pattern indicates that LLMs may inadvertently perpetuate gender stereotypes by associating higher levels of leadership with males, underscoring the need for more sophisticated mechanisms to identify and correct such biases.

These burgeoning and varied endeavors aimed at achieving fairness in LLMs~\cite{chu2024fairness,
gupta2024sociodemographic, li2023survey} highlight the necessity for a comprehensive understanding of how different fair LLM methodologies are implemented and understood across diverse studies. Lacking clarity on these correspondences, the design of future fair LLMs can become challenging~\cite{blodgett2020language}. Consequently, there is a pressing need for a systematic tutorial elucidating the recent advancements in fair LLMs. However, although there are several tutorials that address fairness in machine learning algorithms,
~\cite{ghani2023addressing, hajian2016algorithmic, li2021tutorial, saleiro2020dealing}
these primarily focus on fairness in broader machine learning algorithms. There is a noticeable gap in inclusive resources that specifically address fairness within LLMs, distinguishing it from traditional models and discussing recent developments.

Our tutorial aims to bridge this gap by providing an up-to-date and comprehensive review of existing work on fair LLMs. It begins with a general overview of LLMs, followed by an analysis of the sources of bias inherent in their training processes. We then delve into the specific concept of fairness as it applies to LLMs, summarizing the strategies and algorithms employed to assess and enhance fairness. The tutorial also offers practical resources, including toolkits and datasets, that are essential for evaluating bias in LLMs. Furthermore, we explore the unique challenges of fairness in LLMs, such as those presented by word embeddings and the language generation process. Finally, the tutorial concludes by addressing the current research challenges and proposing future directions for this field.

\textbf{Previous tutorial.} To the best of our knowledge, no other tutorial on fairness in LLMs has been presented at CIKM or other similar venues. 

\section{Tutorial Outline}

We plan to give a half-day tutorial (3 hours plus breaks). To ensure our tutorial remains engaging and interactive, we intend to accomplish as follows: \textbf{i) Case Studies Introduction.} We'll start with a series of case studies that highlight specific instances of bias within LLMs. By grounding our discussion in real-world examples, we aim to help contextualize the discussion and make it more relatable for the audience. We aim to encourage participants to share their thoughts on these cases and foster dialogue. \textbf{ii) Interactive Bias Discussion.} An integral part of our tutorial will involve presenting participants with various LLM outputs and prompts. We'll then facilitate a discussion to identify and analyze potential biases within these examples. \textbf{iii) Fair LLMs Discussion.} We will explore strategies and algorithms for developing fairer LLMs through practical examples. Following this, a presentation of useful tools and datasets for assessing fairness in LLMs will take place to provide participants with concrete tools and methodologies for fairness in LLMs. \textbf{iv) Q\&A Discussion.} The tutorial will culminate in a Q\&A session, allowing participants to ask questions and seek clarifications on any aspects of the session. Additionally, we will make tutorial materials, such as the description, presentation slides, and pre-recorded videos, available for post-tutorial access and dissemination.
\vspace{-0.2cm}
\subsection{Agenda}
The outline of the tutorial is as follows:
\begin{itemize}[noitemsep,topsep=0pt]   
    \item Part I: Background on LLMs (\textbf{30 minutes})
        \begin{itemize}
            \item Introduction to LLMs
            \item Training Process of LLMs
            \item Root Causes of Bias in LLMs
        \end{itemize}
        
    \item Part II: Quantifying Bias in LLMs (\textbf{60 minutes})
        \begin{itemize}
            \item Demographic representation~\cite{brown2020language,liang2022holistic,mattern2022understanding}

            \item Stereotypical association~\cite{abid2021persistent, brown2020language, liang2022holistic}

            \item Counterfactual fairness~\cite{li2023fairness,liang2022holistic}

            \item Performance disparities~\cite{liang2022holistic, wan2023biasasker, zhang2023chatgpt}
            
            

        \end{itemize}
        
    \item Part III: Mitigating Bias in LLMs (\textbf{40 minutes})
        \begin{itemize}
            \item Pre-processing~\cite{fatemi2021improving, lester2021power, yogarajan2023tackling}
            \item In-training~\cite{lauscher2021sustainable, park2023never, ravfogel2020null}
            \item Intra-processing~\cite{akyurek2023dune, huang2019reducing, mitchell2021fast}
            \item Post-processing~\cite{dhingra2023queer, kaneko2024evaluating, tokpo2022text}
        \end{itemize}
        
    \item Part IV: Resources for Evaluating Bias (\textbf{30 minutes})
        \begin{itemize}
            \item Toolkits~\cite{bellamy2019ai, google_jigsaw_2017, saleiro2018aequitas}
            \item Datasets~\cite{dhamala2021bold, huang2023trustgpt, levy2021collecting, neveol2022french, rudinger2018gender}
        \end{itemize}
        
    \item Part V: Challenges and Future Directions (\textbf{20 minutes})   
        \begin{itemize} 
            \item Formulating Fairness Notions
            \item Rational Counterfactual Data Augmentation
            \item Balancing Performance and Fairness in LLMs
            \item Fulfilling Multiple Types of Fairness
            \item Developing More and Tailored Datasets
        \end{itemize}
\end{itemize}

\sloppy
\subsection{Content} 

\textbf{Background on LLMs.} We start by providing the audience with fundamental knowledge about LLMs. Next, we briefly explain the key steps required to train LLMs, including 1) data preparation and preprocessing, 2) model selection and configuration, 3) instruction tuning, and 4) alignment with humans. By examining the training process in detail, we identify and discuss three primary sources contributing to bias in LLMs: i) training data bias, ii) embedding bias, and iii) label bias.

\textbf{Quantifying Bias in LLMs.} To evaluate  bias in LLMs, the primary method involves analyzing bias associations in the model's output when responding to input prompts. These evaluations can be conducted through various strategies including demographic representation, stereotypical association, counterfactual fairness, and performance disparities~\cite{doan2024fairnessdefinitionslanguagemodels}.

Demographic representation~\cite{brown2020language,liang2022holistic,mattern2022understanding} evaluation method assesses bias by analyzing the frequency of demographic word references in the text generated by a model in response to a given prompt~\cite{li2023survey}. In this context, bias is defined as a systematic discrepancy in the frequency of mentions of different demographic groups within the generated text.

Stereotypical association~\cite{abid2021persistent, brown2020language, liang2022holistic} method assesses  bias by measuring the disparity in the rates at which different demographic groups are linked to stereotyped terms (\textit{e.g.,} occupations)  in the text generated by the model in response to a given prompt~\cite{liang2022holistic}. In this context, bias is defined as a systematic discrepancy in the model's associations between demographic groups and specific stereotypes, which reflects societal prejudices. 

Counterfactual fairness~\cite{li2023fairness,liang2022holistic} evaluates bias by replacing terms characterizing demographic identity in the prompts and then observing whether the model's responses remain invariant~\cite{li2023survey}. Bias in this context is defined as the model's sensitivity to demographic-specific terms, measuring how changes to these terms affect its output. 

Performance disparities~\cite{liang2022holistic, wan2023biasasker, zhang2023chatgpt} method assesses bias by measuring the differences in model performance across various demographic groups on downstream tasks. Bias in this context is defined as the systematic variation in accuracy or other performance metrics when the model is applied to tasks involving different demographic groups.

\textbf{Mitigating Bias in LLMs.} We systematically categorize bias mitigation algorithms based on their intervention stage within the processing pipeline.

Pre-processing methods change the data given to the model, like training data and prompts. They do this by using methods like data augmentation~\cite{yogarajan2023tackling} and prompt tuning~\cite{fatemi2021improving, lester2021power}.

In-training methods aim to alter the training process to minimize bias. This includes making modifications to the optimization process by adjusting the loss function~\cite{park2023never} and incorporating auxiliary modules~\cite{lauscher2021sustainable, ravfogel2020null}.

Intra-processing methods mitigate bias in pre-trained or fine-tuned models during inference without additional training. This technique includes a range of methods, such as model editing~\cite{akyurek2023dune, mitchell2021fast} and decoding modification~\cite{huang2019reducing}.

Post-processing methods modify the results generated by the model to reduce biases, which is crucial for closed-source LLMs where direct modification is limited. We use methods such as chain-of-thought~\cite{dhingra2023queer, kaneko2024evaluating} and rewriting~\cite{tokpo2022text} as illustrative approaches to convey this concept.

\textbf{Resource for Evaluating Bias.} In this part, we introduce existing resources for evaluating bias in LLMs. First, we present three essential tools: Perspective API~\cite{google_jigsaw_2017}, developed by Google Jigsaw, detects toxicity in text; AI Fairness 360 (AIF360)~\cite{bellamy2019ai}, an open-source toolkit with various algorithms and tools; and Aequitas~\cite{saleiro2018aequitas}, another open-source toolkit, audits fairness and bias in LLMs, aiding data scientists and policymakers. 

Next, we summarize worth-noting datasets referenced in the literature, categorized into probability-based and generation-based. Probability-based datasets, like WinoBias~\cite{rudinger2018gender}, BUG~\cite{levy2021collecting}, and CrowS-Pairs~\cite{neveol2022french}, use template-based formats or counterfactual-based sentences. Generation-based datasets, such as RealToxicityPrompts~\cite{huang2023trustgpt} and BOLD \cite{dhamala2021bold}, specify the first few words of a sentence and require a continuation. Besides, we will introduce TabLLM ~\cite{hegselmann2023tabllm}, a general framework to leverage LLMs for the classification of tabular data. That approach aims to address the challenge of using LLMs on structured tabular datasets, which are used in high-stakes domains for classification tasks. 

\textbf{Challenges and future directions.} 
The tutorial concludes by exploring open research problems and future directions. Firstly, we discuss the challenges of ensuring fairness in LLMs. Defining fairness in LLMs is complex due to diverse forms of discrimination requiring tailored approaches to quantify bias, where definitions can conflict. Rational counterfactual data augmentation, a technique to mitigate bias, often produces inconsistent data quality and unnatural sentences, necessitating more sophisticated strategies. In addition, balancing performance and fairness involves adjusting the loss function with fairness constraints, but finding the optimal trade-off is challenging due to high costs and manual tuning.

For future directions, it is imperative to address multiple types of fairness concurrently, as bias in any form is undesirable. . 
Additionally, there is a pressing need for more tailored benchmark datasets, as current datasets follow a template-based methodology that may not accurately reflect various forms of bias. 

\section{Target audience and prerequisites for the tutorial}

The tutorial is designed for researchers and practitioners in data mining, artificial intelligence, social science and other interdisciplinary areas, aiming to cater to individuals with varying degrees of expertise. The prerequisites include basic knowledge of probability, linear algebra, and machine learning, while prior knowledge of algorithmic fairness or specific algorithms is not a prerequisite, ensuring accessibility to beginners. This tutorial is designed for 40\% novice, 30\% intermediate, and 30\% expert in order to achieve a good balance between the introductory and advanced materials. To foster a dynamic and participatory learning environment, the tutorial will intersperse lectures with discussion sessions, encouraging attendees to engage, ask questions, and share insights. Furthermore, to extend the tutorial's reach and impact, all materials, ranging from descriptions and slides to pre-recorded videos, will be available for post-tutorial access, supporting continued education and exploration of fairness in LLMs across diverse audiences.

\sloppy

\section{Tutors’ short bio and expertise related to the tutorial}

\noindent \textbf{Thang Viet Doan} is a Ph.D. student in the Knight Foundation School of Computing and Information Sciences at Florida International University. He holds a Bachelor's degree in Computer Science from Hanoi University of Science and Technology (HUST). His current research interests are mainly focused on detecting and mitigating social bias in natural language systems. 

\bigskip

\noindent \textbf{Zichong Wang} is currently pursuing his Ph.D. in the Knight Foundation School of Computing and Information Sciences at Florida International University. His research is centered on mitigating inadvertent disparities resulting from the interaction of algorithms, data, and human decisions in policy development. His work has been honored with the Best Paper Award at FAccT’23 and is a candidate for the Best Paper Award at ICDM’23. Additionally, he actively contributes as a member of the Program Committee/Reviewers for esteemed conferences and journals, including KDD, IJCAI, ICML, ICLR, FAccT, ECML-PKDD, ECAI, PAKDD, Machine Learning, and Information Sciences.

\bigskip

\noindent \textbf{Minh Nhat Hoang Nguyen} is a Ph.D. student at the Knight Foundation School of Computing and Information Sciences, Florida International University. He earned his Bachelor's degree in Data Science and Artificial Intelligence from Hanoi University of Science and Technology (HUST). His research focuses on detecting potential bias in machine learning algorithms, data quality and applying bias mitigation handling methods to deliver fairness in social application.

\bigskip

\noindent \textbf{Wenbin Zhang} is an Assistant Professor in the Knight Foundation School of Computing and Information Sciences at Florida International University, and an Associate Member at the Te Ipu o te Mahara Artificial Intelligence Institute. His research investigates the theoretical foundations of machine learning with a focus on societal impact and welfare. In addition, he has worked in a number of application areas, highlighted by work on healthcare, digital forensics, geophysics, energy, transportation, forestry, and finance. He is a recipient of best paper awards/candidates at FAccT’23, ICDM’23, DAMI, and ICDM’21, as well as the NSF CRII Award and recognition in the AAAI’24 New Faculty Highlights. He also regularly serves in the organizing committees across computer science and interdisciplinary venues, most recently Travel Award Chair at AAAI'24, Volunteer Chair at WSDM’24 and Student Program Chair at AIES’23.

\section{Potential Societal Impacts}

This tutorial possesses significant potential for positive societal impacts: i) By illuminating the nuances of fairness in LLMs, it endeavors to ignite research interest and catalyze efforts aimed at advancing fairness within this domain. Given the early stages of current initiatives addressing fairness in LLMs, this tutorial stands as a pivotal milestone in galvanizing further exploration and innovation in the field. ii) Through the exploration of new challenges that remain unaddressed in existing literature, this tutorial has the potential to inspire innovative approaches within the realm of LLMs fairness. By shedding light on these issues, it aims to stimulate critical discourse and foster the development of comprehensive solutions that address the complexities inherent in ensuring fairness within LLMs. iii) In addition to addressing fairness issues, this tutorial emphasizes the importance of developing new datasets that reflect diverse and representative forms of bias. By highlighting gaps in current datasets, it encourages the creation of new ones, aiming to support more accurate and equitable LLM training processes. iv) Beyond its immediate focus on fairness in LLMs, this tutorial endeavors to extend its impact on related research topics by uncovering new problems and elucidating their interconnectedness with fairness considerations. By identifying emerging issues, it seeks to foster interdisciplinary collaboration and facilitate holistic advancements in understanding and addressing societal concerns surrounding LLMs, thus contributing to broader societal progress and well-being.

\section*{Acknowledgement}

This work was supported in part by the National Science Foundation (NSF) under Grant No. 2245895.

\bibliographystyle{ACM-Reference-Format}
\bibliography{reference}

\end{document}